\documentclass[12pt,preprint,review]{elsarticle}

\usepackage{amsmath}
\usepackage{amssymb}
\usepackage{bm}
\usepackage{url}
\usepackage[colorlinks=true]{hyperref}
\usepackage{multirow}
\usepackage{booktabs}
\usepackage{multirow}
\usepackage[figuresright]{rotating}
\usepackage{oubraces}
\usepackage{mathtools}
\usepackage{algorithm}
\usepackage[noend]{algpseudocode}
\usepackage{mathrsfs}
\usepackage{subcaption}
\usepackage{tikz}
\usepackage{pgfplots}

\DeclareMathSymbol{\R}{\mathalpha}{AMSb}{"52}
\DeclareMathOperator*{\E}{\mathbb{E}}
\graphicspath{{images/}}

\newcommand{\addReviewer}[2]{
  \expandafter\newcommand\csname #1\endcsname[1]{{\bf \color{#2} \capitalisewords{#1}:\,##1}}
  \expandafter\newcommand\csname #1cor\endcsname[2]{{\color{#2} \capitalisewords{#1}:\,\st{##1}{\bf ##2}}}
  \expandafter\newcommand\csname #1color\endcsname{#2}
}

\addReviewer{jary}{orange}

\journal{Neurocomputing}

\begin{document}

\begin{frontmatter}

\title{Efficient Continual Learning in Neural Networks with Embedding Regularization}

\author[sapienza]{Jary Pomponi}

\author[sapienza]{Simone Scardapane\corref{cor1}}
\ead{simone.scardapane@uniroma1.it}
\cortext[cor1]{Corresponding author. Phone: +39 06 44585495.}

\author[bologna]{Vincenzo Lomonaco}
\ead{vincenzo.lomonaco@unibo.it}

\author[sapienza]{Aurelio Uncini}

\address[bologna]{Department of Computer Science and Engineering, University of Bologna, Italy}

\address[sapienza]{Department of Information Engineering, Electronics and Telecommunications (DIET), Sapienza University of Rome, Italy}

\begin{abstract}
Continual learning of deep neural networks is a key requirement for scaling them up to more complex applicative scenarios and for achieving real lifelong learning of these architectures. Previous approaches to the problem have considered either the progressive increase in the size of the networks, or have tried to regularize the network behavior to equalize it with respect to previously observed tasks. In the latter case, it is essential to understand what type of information best represents this past behavior. Common techniques include regularizing the past outputs, gradients, or individual weights. In this work, we propose a new, relatively simple and efficient method to perform continual learning by regularizing instead the network internal embeddings. To make the approach scalable, we also propose a dynamic sampling strategy to reduce the memory footprint of the required external storage. We show that our method performs favorably with respect to state-of-the-art approaches in the literature, while requiring significantly less space in memory and computational time. In addition, inspired by recent works, we evaluate the impact of selecting a more flexible model for the activation functions inside the network, evaluating the impact of catastrophic forgetting on the activation functions themselves. 
\end{abstract}


\end{frontmatter}

\section{Introduction}
\label{sec:introduction}

While deep networks have achieved remarkable successes over the last decade, most of their applications are limited to single, isolated problems, where each network has to be redesigned (and re-trained) from scratch. As a result, their training remains daunting in situations where data is scarce and/or computation is expensive. In these scenarios, efficient transfer of information from one task to another (e.g., in reinforcement learning contexts) is widely acknowledged to be a fundamental technique for their deployment \citep{parisi2019continual}. Depending on the number of tasks and their similarity, different applicative scenarios can be devised, ranging from simple transfer learning from one source domain to a target domain (e.g., as is popular today with pre-trained image classification networks), up to full, \textit{lifelong} learning of a single network on a possibly infinite number of interrelated tasks.

Artificial general intelligence must be able to learn and remember many different tasks \citep{definitionOfML} and this is particularly complex in real-world scenarios. The ability of an agent to learn consecutive tasks without forgetting how to solve previously learned tasks is called continual learning (CL). CL is a property difficult to obtain, because when the network learns how to solve a task the information about how to solve previously learned tasks is generally lost if no countermeasure is adopted; this phenomenon is called \textit{catastrophic forgetting} \citep{FRENCH1999128}, and it occurs when the network, trained on a task, changes its weights in order to meet the objectives of the new task. As a result, the network accuracy on a previous task is drastically reduced after a few training updates on successive tasks. Whilst recent results on deep networks have improved the results on a wide variety of domains, relatively less progress has been made in reaching continual learning on artificial neural networks. 

Because of its importance, many ideas and algorithms were proposed recently for an effective CL of neural architectures. One class of algorithms, known as progressive training algorithms \citep{chen2015net2net,rusu2016progressive,schwarz2018progress}, incrementally builds the structure of the network for each task being processed, trying to re-use as much as possible of the previous architecture in the process. Another class of algorithms, known as \textit{rehearsal methods} \citep{robins1995catastrophic,shin2017continual,atkinson2018pseudo}, save the information about past tasks in an external memory, and continually retrain the network on this memory in order to maintain its performance. Finally, a third class of algorithms try to reuse a single neural network by instead including multiple \textit{regularization} penalties \citep{kirkpatrick2017overcoming} to stabilize the behaviour of the network with respect to previous tasks. Generally speaking, rehearsal and progressive strategies tend to perform well but scale poorly with respect to the number of tasks, and might have a large computational complexity. Differently, regularization strategies are simple to implement, tend to require less memory, but their performance might not match naive rehearsal.

One key problem when implementing a regularization strategy is deciding what information best represents the behaviour of the network and, as a result, what form the regularization penalty should take. Proposals in this sense range from regularizing individual weights based on their importance \citep{kirkpatrick2017overcoming}, to specific gradients \citep{lopez2017gradient}, outputs \citep{li2017learning} (similar to the idea of knowledge distillation), using information from an external memory or a combination of these. 
Some state-of-the-art algorithms in this sense are described in-depth in Section \ref{sec:continual_learning_in_neural_networks}. Choosing the proper level of abstraction in this sense is not trivial, and the choice highly influences the level of accuracy of these techniques, and their ability to scale on multiple tasks.

\subsection*{Contributions of this paper}

In this paper, we propose a novel regularization-rehearsal approach for continual learning. 
Our insight is that an efficient way to regularize the behaviour of the network is to act on its internal \textit{embeddings}, i.e., the activations of one or more layers close to the exit. We argue that regularizing the embeddings provide more information as opposed to simply asking for similar outputs, allowing the network more degrees of freedom in choosing how best to enforce the penalty, while being less stringent than asking for specific, individual weights to be preserved.

Because our method requires a small external memory of past activations to work efficiently, we also devise a simple sampling strategy for this memory to maximize the information provided by the regularization term, i.e., we try to sample maximally those elements in memory that can provide the strongest regularization on the model.

In addition to the previous contributions, we also evaluate the impact of more flexible activation functions, namely, kernel activation functions (KAFs, \citep{SCARDAPANE201919}), on the CL scenario. Several papers \citep{goodfellow2013maxout,agostinelli2014learning,jin2016deep,qian2018adaptive,SCARDAPANE201919} recently have shown that, in the single-task scenario, training the activation functions can provide significant improvements in performance and in the number of layers required to solve the given task. The intuition is that giving more freedom to the activation functions inside the network, whenever these are properly normalized, can have a beneficial impact on the learning of the networks. However, very little has been done to evaluate whether these benefits extend to the CL case or if, on the opposite, flexible activation functions can suffer from catastrophic forgetting themselves. 

To this end, in the experimental section we compare our proposed regularization technique with multiple state-of-the-art regularization algorithms for CL, both on standard neural networks and on neural networks endowed with KAFs, hereafter denoted as \textit{Kafnets}.

\subsection*{Organization of this paper}

The rest of the paper is structured as follows. Section \ref{sec:continual_learning_in_neural_networks} describes the problem of CL in neural networks and some 
techniques for performing CL through proper regularization. Then, after elaborating on the advantages and shortcomings of each of these techniques, in Section \ref{sec:proposed_embedding_regularization_for_continual_learning} we propose a novel method for performing the regularization, that we call embedding regularization (ER). We provide a comprehensive experimental evaluation of the proposed ER algorithm in Sections \ref{sec:experimental_setup} and \ref{sec:experimental_evaluation}, before giving some concluding remarks in Section \ref{sec:conclusions}.

\section{Continual learning in neural networks}
\label{sec:continual_learning_in_neural_networks}

\subsection{Problem formulation}

We follow the formulation of continual learning from \citep{lopez2017gradient}. We receive a (possibly infinite) sequence of labeled examples, and we suppose that  each example is described by a triplet $(x, t, y) \in \mathcal{X} \times \mathbb{N^+} \times \mathcal{Y}$, where $x$ is an input (e.g., an image or a vector of features), $t$ is an integer identifying a particular task, and $y$ is the corresponding label of $x$ (e.g., the presence of a specific object inside the image). 

In this formulation, $x$ can belong to any domain (e.g., an image, or a vector in a suitable Euclidean space), although most benchmarks in the CL literature have considered the image domain \cite{parisi2019continual}. As a motivating example, images with $t=0$ could contain either cars or trucks (two classes), while images with $t=1$ could contain planes, helicopters, or boats (three classes) and so on. Crucially, we assume that a pair $(x, y)$ is locally i.i.d. given $t$, i.e., it is sampled from some probability distribution $P_{t}$ that fully specifies the task. The total number of tasks can be known \textit{a priori} or not. 

In addition, several works in the literature consider sequential scenarios where a full dataset belonging to a given task is received and processed (even with multiple passes over it), before switching to the next task. Nonetheless, methods developed for the sequential case can generally be extended to the non-sequential (online) case, and \textit{vice versa}.

In continual learning we want to design a training algorithm for a neural network $f_\theta(x, t)$, parameterized by a vector $\theta$, such that $f_\theta(x, t) \approx y$ for any possible task and example. Because we assume potential correlations between different tasks, the network should share as much as possible the information contained in $\theta$ across all possible problems.

For a single task $t$ (e.g., only images of cars/trucks), we could find the optimal $\theta$ by minimizing the expected risk of the network as:

\begin{equation}
    \theta^* = \arg\min_{\theta} \left\{ \displaystyle \E_{(x, y) \sim P_1} \mathcal{L} \left(f_\theta(x, t), y\right) \right\} \,,
    \label{eq:single_task_learning}
\end{equation}

\noindent where $\mathcal{L}(\cdot, \cdot)$ is a loss function and the expectation is taken with respect to all possible data of the single task. If we are provided with a dataset of $B$ examples $\left\{x_i, t, y_i\right\}_{i=1}^B$ from the task, then \eqref{eq:single_task_learning} can be approximated by minimizing:

\begin{equation}
    J_t(\theta) =  \left\{ \displaystyle \frac{1}{B} \sum_{i=1}^B \mathcal{L}\left(f_\theta(x_i, t), y_i \right) \right\} \,.
    \label{eq:single_task_learning_empirical}
\end{equation}

\noindent Practically, \eqref{eq:single_task_learning_empirical} can be solved by taking mini-batches of data and using these to compute approximate gradients of the cost function $J(\theta)$ \citep{bottou2018optimization}. 

Extending this formulation to more than a single task is non-trivial for the problems described in Section \ref{sec:introduction}, mostly because of the  catastrophic forgetting of previously learned information \citep{parisi2019continual}. In particular, suppose we are provided with a separate dataset of $B'$ examples $\left\{x_i, t', y_i\right\}_{i=B+1}^{B+B'}$ from a different task $t'$ (e.g., a new mini-batch of data containing planes and similar objects). A naive solution would be to solve a problem $J_{t'}(\theta)$ similar to \eqref{eq:single_task_learning_empirical}, starting however from the previous solution $\theta^*$. Catastrophic forgetting then occurs whenever the solution of this new optimization problem performs poorly on the original task. 

Next, we describe many state-of-the-art regularization solutions for approaching this problem. Note how, as the size of the datasets shrink, and as we iterate this process, we are back at the sequential scenario described at the beginning of this section. Nonetheless, it is easier to describe continual learning techniques in this simplified context.

\subsection{Regularization by weight consolidation}

The idea of elastic weight consolidation (EWC) \citep{kirkpatrick2017overcoming} is to impose a regularization term on successive tasks to preserve as much as possible the weights in $\theta$ that were essential to the previous task. This is achieved by augmenting $J_{t'}(\theta)$ with a quadratic penalty as follows:

\begin{equation}
    R(\theta) = \frac{\lambda}{2} \cdot \sum_{z} F_z \left( \theta_z^* - \theta_z \right)^2 \,,
    \label{formula:ewc_reg}
\end{equation}

\noindent where $\lambda$ is a hyper-parameter, $\theta^*$ is the weight vector after training on the first task, and $F_z$ represents the $z$th diagonal element of the Fisher information matrix:

\begin{equation}
    F_z = \frac{1}{B} \sum_{i=1}^B \frac{\partial^2 \log\left[ f_{\theta^*}(x_i, t) \right]}{\partial \theta^*_z} \,.
\end{equation}

\noindent EWC can be justified from a Bayesian perspective by assuming that the network posterior distribution can be fully specified by the mean vector $\theta^*$ and a diagonal precision matrix given by the elements $\left\{F_z\right\}_{z}$. When training on more than two tasks, different penalties can be combined by storing the mean vectors and Fisher diagonal matrices for each task \citep{kirkpatrick2017overcoming}, or a single composite penalty can be obtained in various ways \citep{liu2018rotate,huszar2018note,kirkpatrick2018reply}.

More in general, when faced with a sequential scenario, we would generate a sequence of weight vectors $\theta^1, \theta^2, \ldots$ and corresponding diagonal Fisher matrices $F^1, F^2, \ldots$. At a generic iteration $n$, a simple online extension of EWC \citep{schwarz2018progress} computes a single dynamic penalty by keeping the last mean vector $\theta^{n-1}$ and updating the Fisher information matrix as:

\begin{equation}
    F^n = F^n + \gamma F^{n-1} \,,
\end{equation}

\noindent where $\gamma \in \left[0, 1\right]$ is a discounting factor. In the online version of the approach only the optimal weights from the last task are maintained into memory. Alternatively, one can keep an external memory $\mathcal{M}$ of samples from previous tasks, and use this memory to build an aggregated Fisher information matrix for every iteration. However, how best to sample from previous tasks remains a significant open challenge \citep{kirkpatrick2017overcoming}.

\subsection{Regularization by knowledge distillation}

While EWC has been shown to work well in some scenarios, regularizing individual weights might not always work optimally, especially when different penalties start to exert conflicting influences. Learning without forgetting (LWF) \citep{li2017learning} is another method for continual learning whose idea is to simply preserve the outputs of a network on older tasks, while leaving it free to modify weights as necessary to achieve this objective. To do so, LWF employs a variant of knowledge distillation \citep{hinton2015distilling}.

In our notation, $f_{\theta^*}(x, t')$ is the output of the network using the previous weights (at iteration $n-1$) on an example from the separate task $t'$. Suppose we are in a binary classification task, in which case the output is the probability of $x$ belonging to the class specified by $t'$. To regularize the network, we add a penalty term to $J_{t'}(\theta)$ forcing the new probabilities to be similar to the old probabilities using a cross-entropy term:

\begin{equation}
    R(\theta) = - \lambda \cdot \frac{1}{B'} \sum_{i=B+1}^{B'} f_{\theta^*}(x_i, t') \log\left[ f_{\theta}(x_i, t') \right] \,,
\end{equation}

\noindent where $\lambda$ has the same meaning as in the previous section. For regression problems one can use the mean-squared error, while for multi-class classification problems it is possible to improve on this basic scheme by re-weighting the probabilities using a new temperature hyper-parameter \citep{li2017learning}. LWF has the advantage that the training process can potentially discover new weight configurations able to preserve older behaviors.

\subsection{Synaptic Intelligence}

Synaptic Intelligence (SI), introduced in \cite{SI}, is a variant of EWC. The main idea is to calculate the weight importance in an online fashion, instead of calculating the Fisher matrix.
To do it the authors characterized the trajectory of the network parameters over the optimization process to evaluate the importance of a parameter's change in the loss' decrease. In particular, a weight importance w.r.t. optimization can be defined as: 

\begin{equation}
    F_z = \frac{\Delta J_z}{(\Delta_z)^2 + \xi}
\end{equation}
\noindent where $\Delta J_z$ is the sum of instantaneous changes to the parameter $\theta_z$ along the optimization trajectory of $J_t$ (see \cite{SI} for the full definition), $\Delta_z$ is its absolute variation from the starting condition, and $\xi$ a small constant to bound the expression in case of numerical issues.

The main advantage of this approach is the possibility to calculate $F_z$ using the information that are available during the training and no extra computation is needed. The magnitude of $F$ can be scaled by proper setting of a scale factor $1 \ge c \ge 0$, which regularizes how much of the past information needs to be saved. This factor can then be used in a EWC-like regularization term replacing the Fisher information matrix.
    
\subsection{Gradient episodic memory}

Gradient episodic memory (GEM) \citep{lopez2017gradient} is another technique which regularizes a model by working neither on individual weights nor on the outputs, but on the \textit{gradients} of the model. In GEM, a subset of past tasks is saved into an external fixed size memory, and used to constrain the current gradients to avoid the increasing of losses associated to past tasks, but at the same time allowing their decreasing; this could lead to positive transfer of knowledge to past tasks.

To this end, denote by $g_t$ the gradient of the loss obtained with a sample from the past task $t$ using the current network, and by $g_{t'}$ the gradient vector on the current task, the method alleviates catastrophic forgetting by enforcing the gradients to point in the same direction: 
\begin{align*}
 \langle g_{t'}, g_t \rangle  \ge 0 \,,
\end{align*}
\noindent if the inequality constraints are satisfied, then the proposed parameter update is unlikely to increase the loss on previous tasks. In this case the past information is preserved. Otherwise there is at least one previous task that would experience an increase in loss after the parameter update.


In order to enforce the previous penalty, at every optimization step a quadratic program with inequality constrains is solved: 
\begin{align*}
 \underset{v}{\text{minimize}} & \quad \frac{1}{2} v^T G_t G_t^T v + g_{t'}^T G_t^T v \\
 \text{subject \ to}  & \quad v \ge 0
\end{align*}

\noindent where $G_t$ is a matrix containing all the gradients associated to the past task samples (i.e., each row is the gradient associated to one of the $B$ samples). The new gradients are calculated as $\bar{g} = G^T_t v^* + g_{t'}$, where $v^*$ is the solution to the QP. Basically, the method forces the new gradients to go in the same direction of the gradients associated to past tasks. This method requires more memory than a simpler regularization approach such as EWC, and more computational time, but could provide improved performance. 


Recently a new version of GEM, called A-GEM, has been proposed in \citep{AGEM}. This approach tries to ensure, using a different QP w.r.t. GEM, that at every training step the average episodic memory loss over the previous tasks does not increase; the resulting QP used in A-GEM is faster and easier to minimize. The main difference between A-GEM and GEM is that GEM has better guarantees in terms of worst-case forgetting of each task since it prohibits an increase of any past task specific loss, while A-GEM has better guarantees in term of average accuracy obtained, since GEM can prevent a gradient update on the current task in order to satisfy the constraints. 


\section{Proposed embedding regularization (ER) for continual learning}
\label{sec:proposed_embedding_regularization_for_continual_learning}

All methods described in the previous section are similar, in the sense that they try to preserve the behaviour of the neural network on previous tasks by modifying the (unconstrained) descent direction in a direction that alleviates catastrophic forgetting. They differ, however, in what information is the best representation of such behavior: individual weights for EWC and SI, gradient information for GEM, or actual outputs for LWF.

As we stated in Section \ref{sec:introduction}, our aim in this paper is to complement these methods with a separate regularization technique acting on the \textit{embeddings} of the neural network, i.e., the activations of the second-to-last layer of the model. We argue this is both more flexible than individual weights, and more informative than the outputs themselves, as evidenced by the large literature on using pre-trained networks for a variety of fine-tuning tasks and few-shot learning \citep{oquab2014learning}. The method we propose is connected to the less-forgetful learning (LFL) \citep{jung2018less} method in domain adaptation, and in a lesser form to neural graph machines \citep{bui2017neural} for graph-based learning.
LFL is not suitable for a continual learning scenario, since the idea is that, given a source network trained on a source domain, the weights of the source network are used as the initial weights of the target network and then the parameters associated to the softmax layer are frozen; this network is then trained. However, while LFL also regularizes a network on a new domain using embeddings, it is not designed to work on more than two tasks, and its memory requirements scale linearly with the number of tasks, since it requires a network for each one. Also, the features regularization is done by minimizing the distance between the features associated to the new task (e.g., the target domain) extracted using the old network and the new one; so the domains need to be correlated somehow, also because the number of classes remains unchanged from one network to another and the softmax layer is fixed. Conversely, we are interested in the more general CL scenario described above, and in designing a method with a fixed, low-memory overhead.


\subsection{General description of the proposed method}

\begin{figure}
    \centering
    \includegraphics[width=0.9\columnwidth]{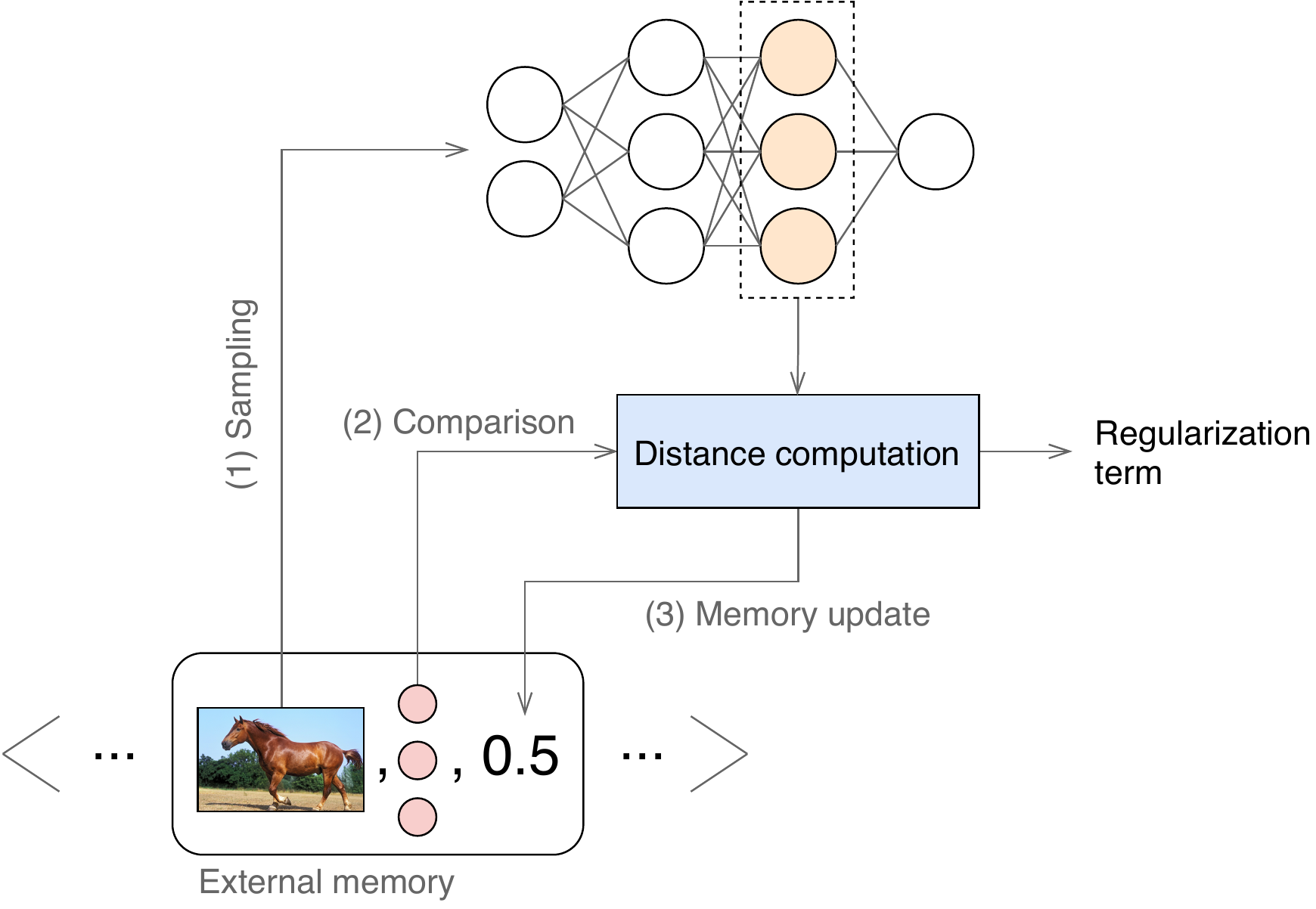}
    \caption{Overall schema of the proposed framework. See the text for a description of the different steps.}
    \label{fig:embedding_regularization}
\end{figure}

The overall schema of the proposed embedding regularization (ER) technique is provided in Fig. \ref{fig:embedding_regularization}. Similarly to before, we represent the continual learning scenario with a small external memory $\mathcal{M} = \left\{ (x_m, t_m, h_m, p_m) \right\}_{m=1}^M$, whose size $M$ is selected by the user, and where each element of the memory is a quadruplet containing:

\begin{itemize}
    \item An image $x_m$ that the network processed previously and the corresponding task $t_m$;
    \item The embedding $h_m$ corresponding to the activation of the second-to-last layer of $f_\theta(x_m, t_m)$, computed when the network was trained last on task $t_m$;
    \item A sampling weight $p_m$. The set of weights $\left\{p_m\right\}_{m=1}^M$ should define a probability distribution over the elements of $\mathcal{M}$.
\end{itemize}

\noindent Whenever we want to regularize our model, we sample an element $(x_m, t_m, h_m)$ from the past task memory $\mathcal{M}$ (or a mini-batch of elements) with a probability proportional to $p_m$. We then compare the embedding with the current embedding $\hat{h}_m$ computed with the current weights of the network:

\begin{equation}
    R(\theta) = \lambda \cdot d\left(h_m, \hat{h}_m \right) \,,
\end{equation}

\noindent where we use the cosine similarity between the two vectors as distance function $d(\cdot, \cdot)$ in our implementation. The overall algorithm is summarized in Algorithm \ref{alg:emb}, while in the following we elaborate more in-depth with respect to the sampling strategies we can adopt.

        \begin{algorithm}[t]
        \caption{Embedding regularization (ER)}\label{euclid}
        \label{alg:emb}
        \begin{algorithmic}[1]
        \Procedure{Train Procedure}{}
        \State \textit{Given}: a neural network $f$ with current weights $\theta$
        \State \textit{Given}: a mini-batch of $B'$ elements from a task $t$.
        \State \textbf{Step 1} Perform one step of optimization on the mini-batch.
        \State \textbf{Step 2} Regularize the model with EmbeddingRegularization().
        \State \textit{When the task is over:} Update the external memory $\mathcal{M}$.
        \EndProcedure
        
        \Procedure{EmbeddingRegularization}{}
        \State Sample $m$ values $\left\{ (x_i, t_i, h_i, p_i) \right\}$ from $\mathcal{M}$

        \State $m \leftarrow \lambda * \text{mean} \left(\{d(h_i, \hat{h}_i )\}_{\ i=1 \dots m}\right)$
        
        \State Perform one step of optimization on $m$.
        \EndProcedure

        \end{algorithmic}
        \end{algorithm}

\subsection{Selecting and updating the sampling probabilities}
\label{subsec:selecting_and_updating_the_sampling_probabilities}

A correct sampling from the external memory is the key of this method, since sampling unrepresentative images might lead to catastrophic forgetting. The simplest way to do this, besides using uniform sampling probabilities for each value in $\mathcal{M}$, is to count how many times a sample $m_i \in \mathcal{M}$ has already been chosen and assign to the weight a value equal to the inverse of how many times it has been extracted (in the following, this is referred to as the \textit{frequency} sampling strategy). In this way, however, the weights are not a real indicator of the importance of $m_i$, given the current task. 

Another way to do that is to assign an initial value to each weight and then, when $m_i$ is picked and the distance $d_i = d\left(h_i, \hat{h}_i \right)$ calculated, assign a weight $p_i$ proportional to the distance $d_i$ (\textit{distance} sampling strategy). In this way any image that is associated to a high distance will be drawn more frequently, because it contains some information that is in contrast to the current task and needs to be enforced multiple times. 

Moreover, the distance between the images can be used, by using the structural distance from an image in the memory and some images from the current task. When a new batch is encountered, an external network, whose parameters are randomly initialized and not trained, is used to extract the features from the images present in the current batch and from the images drawn from the memory (\textit{pretrained reference} sampling strategy). Then the weight $p_i$ is set equal to the mean distance between the image $x_i$ from the memory and all the images in the current batch. The idea is to use a distance that won't change during the training and was inspired by \citep{ulyanovDeepPrior}.

\section{Experimental setup}
\label{sec:experimental_setup}

\subsection{Datasets and metrics}
For evaluating the proposed method, we consider the following datasets:
\begin{itemize}
    \item Permuted MNIST: \citep{kirkpatrick2017overcoming} it is a variant of the popular MNIST dataset of handwritten digits \citep{minst}, in which each task is a random fixed permutation of the pixels and the images are treated as vectors of features (flattened images). In this experiments the output of the network is fixed to 10. This is a relatively easy benchmark, allowing to easily compare with other techniques. In our experiment the number of tasks is set to 4.
    
    \item CIFAR10: another dataset widely used in the literature. It consists in 10 classes with 60000 colour images each. To adapt this dataset for a continual learning scenario the classes are grouped in $n$ sets, in our case $ n = 5$; so each task contains $2$ consecutive classes. In this case the scenario is the CL one. 
    \item CIFAR100: a variant of the CIFAR10 dataset with 100 classes. The number of images are the same as CIFAR10 but, since the classes are 100, each class has less images and this makes the training harder. We grouped the classes into $10$ tasks.
\end{itemize}
     
To evaluate the efficiency and to compare the methods different metrics have been used. All these metrics are calculated on a matrix $R \in \mathbb{R}^{N \times N}$, where N is the number of tasks, and each entry $R_{ij}$ contains the test classification accuracy on task $j$ when the training on task $i$ is completed. The metrics used are the following.

\textbf{Accuracy}: it is the average accuracy on the trained task, it considers the elements of the diagonal as well as the elements below it: 

\begin{align*}
 \text{Accuracy} = \frac{\sum_{i>j}^N R_{i,j}}{\frac{1}{2} N (N+1)} \,.
\end{align*}

\noindent This metric aims to show the averaged performance of the model in every step of the training and for each task.

\textbf{Remembering}: it measures how much information from the old tasks is remembered during the training on the new one. Given a score called backward transfer:

\begin{align*}
 \text{backward transfer} = \frac{\sum_{i=2}^{N} \sum_{j=1}^{i-1} (R_{i,j} - R_{j,j})}{\frac{1}{2} N (N-1)}  
\end{align*}

\noindent that measures the influence that learning a task has on the previously learned tasks, the remembering score is defined as:

$$
\textbf{Remembering} = 1 - |\min(0, \text{backward transfer})| \,.
$$

\noindent Having a score equals to one means not only that the model is remembering everything about the past tasks, but it can also improve these. To measure how the old tasks have been improved the metric used is \textbf{Positive Backward Transfer} = $\max(0, \text{backward transfer})$.

We used these metrics because they embody all the import aspects of CL problems: the ability of a model to being able to classify past tasks, by reducing the CF (\textbf{remembering}), but at the same time the ability to correctly classify all the tasks  during the training phase (\textbf{accuracy}), including the current task. Also, they are jointly important: a model with high accuracy and low remembering is a model not capable of alleviating CF, and will rewrite all the weights at each task; on the other hand low accuracy and high remembering tells us that the constraints applied on the model are too restrictive, blocking the learning of the current task by remembering all the past information. 

These metrics and more are defined in \citep{metrics}. The complete code of our experimental evaluation can be found online.\footnote{ \url{https://github.com/jaryP/Master-thesis}}

\subsection{Architectures and methods}

On the the MNIST experiment a fully-connected neural network with 4 hidden layers of 400 ReLU units is used. This network is trained using SGD with learning rate equal to $1e-3$. For the CIFAR10/100 the CNN is the same used in \citep{zenke2017continual}, except for the CIFAR100 case where the number of kernels is doubled. The CNNs have been trained using the Adam optimizer with learning rate set to $1e-3$; the optimizer state was reset after training on each task.

As stated in the introduction, as a further contribution we also decided to compare the standard NNs to the Kafnets \citep{SCARDAPANE201919}, a class of neural networks using KAFs as activation functions. Briefly, using this architecture each activation function is allowed to change shape during training using a small number of adaptable coefficients, and the aim of this set of comparisons is to explore whether this additional number of degrees of freedom is beneficial or not in terms of CL. Each Kafnet used has the same architecture of the NN counterpart but the size of each layer/kernel is reduced by $30\%$, in order to have roughly the same number of adaptable parameters for both architectures. The hyper-parameters for the KAFs use the same values from \citep{SCARDAPANE201919}. 

To demonstrate the effectiveness of the proposed ER algorithm we compare it with GEM, online EWC, SI, and the full training without any approach to alleviate catastrophic forgetting as further comparisons; we discarded LWF because it showed no good performance in our preliminary experiments on CL (being designed for a single transfer scenario), and also SI from KAFs experiments, given the poor performance obtained on the standard ANN (see below). The memory size for each task in the GEM approach was set to 200 for each task in CIFAR10/MNIST and 500 for each task in CIFAR100. For our method, memory was fixed to a smaller size of $100$ images per task. All of the values were fine-tuned to provide the best accuracy, while in Section \ref{subsec:impact_of_the_memory_size} we evaluate the impact of selecting different values for the external memory, for both GEM and ER. For our proposed method, we use the distance sampling strategy in the following, and evaluate the impact of the other sampling strategies described in Section \ref{subsec:selecting_and_updating_the_sampling_probabilities} in Section \ref{subsec:impact_of_the_sampling_strategy}.

Finally, $\lambda$ was set to 1 in every experiment with the exception of CIFAR100, in which it was set to 5, and the $c$ factor in SI was set to 1 for permuted MNIST and 0.01 for the other tasks. All of the parameters were chosen after a preliminary empirical analysis.


\section{Experimental evaluation}
\label{sec:experimental_evaluation}

\begin{table}[t]
\centering
\resizebox{\textwidth}{!}{%
\begin{tabular}{cc|c|c||c|c||c|c||c|c|}
\cline{3-10}
 &  & \multicolumn{2}{c||}{Perm-MNIST} & \multicolumn{2}{c||}{CIFAR10} & \multicolumn{2}{c||}{CIFAR100} & \multicolumn{2}{c|}{Average} \\ \cline{3-10} 
 &  & Accuracy & Remembering & Accuracy & Remembering & Accuracy & Remembering & Accuracy & Remembering \\ \hline
\multicolumn{1}{|c|}{\multirow{5}{*}{\rotatebox[origin=c]{90}{Standard NN}}} & None & 0.86 & 0.9509 & 0.6548 & 0.6858 & 0.2487  & 0.4789 & 0.5878 & 0.7052 \\ \cline{2-10}
\multicolumn{1}{|c|}{} & EWC & 0.8656 & 0.9894 & 0.6549 & 0.7856 & 0.2187 & 0.5971 & 57.94 & 0.7907 \\ \cline{2-10} 
\multicolumn{1}{|c|}{} & SI & 0.8592 & 0.9373 & 0.655 & 0.6836 & 0.2532 &  0.4693 & 58.91 & 0.68 \\ \cline{2-10}  
\multicolumn{1}{|c|}{} & GEM & \bfseries 0.9421 & \bfseries 1 (+0.0894) & 0.8127 & 0.8817 & 0.3211 & 0.6629 & 0.6919 & \bfseries 0.8511 \\ \cline{2-10}  
\multicolumn{1}{|c|}{} & ER & 0.9128 & 0.9908 & \bfseries 0.8428 & \bfseries 0.8905  & \bfseries 0.3813 & \bfseries 0.6643 & \bfseries 71.23 & 0.8485 \\ \hline \hline
\multicolumn{1}{|c|}{\multirow{4}{*}{\rotatebox[origin=c]{90}{Kafnet}}} & None & 0.8526 & 0.6489 & 0.6620 & 0.6197 & 0.2304 & 0.4441 & 0.5948 & 0.5709 \\ \cline{2-10} 
\multicolumn{1}{|c|}{} & EWC & 0.8816 & 0.9939 & 0.7107 & \bfseries 0.8902 & 0.2030 & 0.6030 & 59.84 & 0.8290 \\ \cline{2-10} 
\multicolumn{1}{|c|}{} & GEM & \bfseries 0.912 & \bfseries 1 (+0.0157) & 0.7320 & 0.8233 & 0.2903 & 0.638 & 64.47 & \bfseries 0.8250 \\ \cline{2-10}  
\multicolumn{1}{|c|}{} & ER & 0.8973 & 0.9937 & \bfseries 0.7856 & 0.8181 & \bfseries 0.3205 & 0.5975 & \bfseries  66.78 & 0.8030 \\ \hline
\end{tabular}%
}\
\caption{Overall results on the three datasets, two architectures, and different strategies for CL. The plus symbol means that the approach is capable not only of remembering completely the past information about the past tasks, but also to improve the classification of it during the training on new tasks. The average column contains the averaged Accuracy and Remembering obtained on all the benchmarks.}
\label{table}
\end{table}

\subsection{Overall comparison of the algorithms }

Table \ref{table} summarizes all the scores obtained on all datasets, methods and architectures. We decided to use the training without any approach to alleviate catastrophic forgetting as baseline. Overall, the ER method performs similarly or better than GEM, for both accuracy and remembering; with the exception of the MNIST dataset, in which GEM performs better. In general, including more flexible activation functions does not seem to bring benefit to the CL task, worsening the results in most scenarios, possibly because of catastrophic forgetting occuring on the activation functions themselves.

\begin{figure}[t]
\begin{subfigure}{\textwidth}
  \centering
  \includegraphics[width=\linewidth]{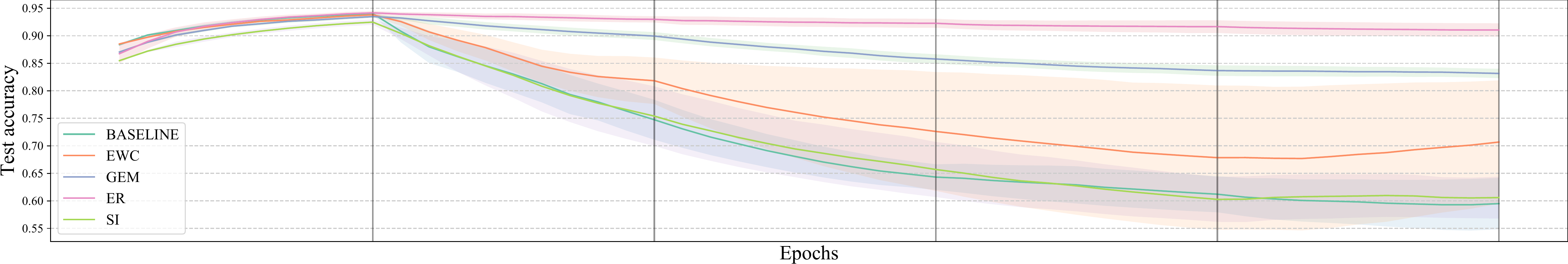}
  \caption{Standard NN}
  \label{cifar10:ann}
\end{subfigure}

\begin{subfigure}{\textwidth}
  \centering
  \includegraphics[width=\linewidth]{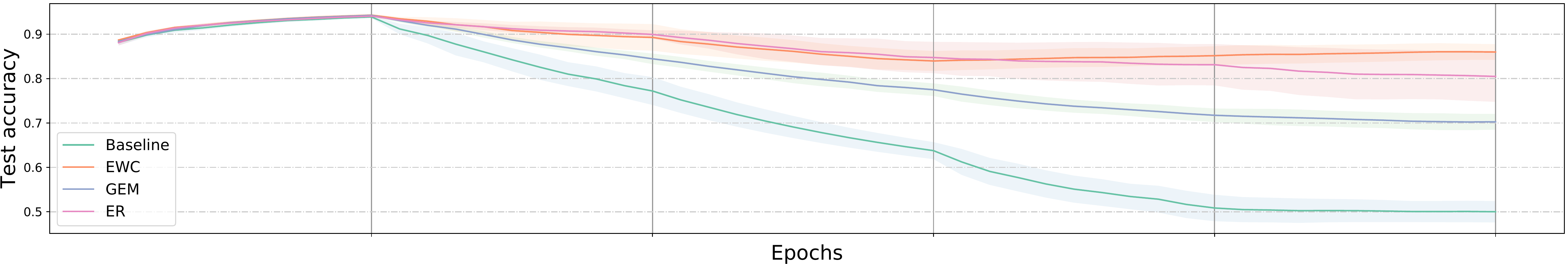}
  \caption{Kafnet}
  \label{cifar10:kaf}
\end{subfigure}

\caption{Evolution of accuracy on task 1 when training on CIFAR10.}
\label{cifar10:task1}
\end{figure}

\begin{figure}[t]
\begin{subfigure}{\textwidth}
  \centering
  \includegraphics[width=\linewidth]{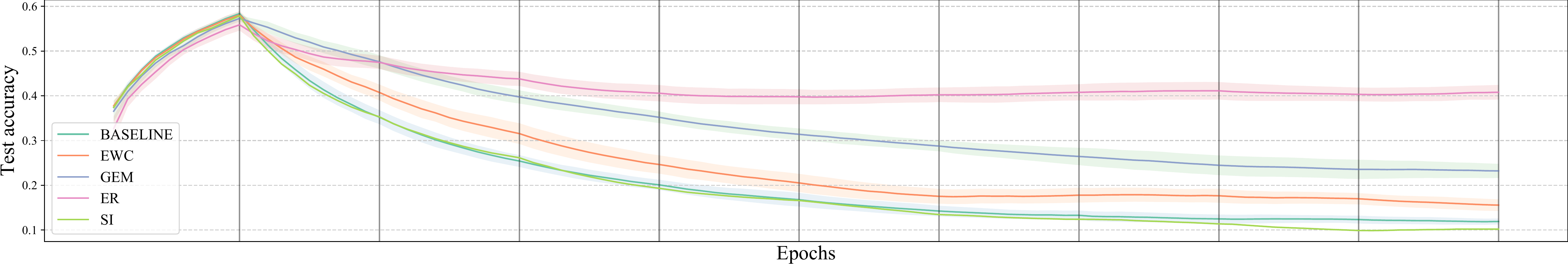}
  \caption{Standard NN}
  \label{cifar100:ann}
\end{subfigure}
\begin{subfigure}{\textwidth}
  \centering
  \includegraphics[width=\linewidth]{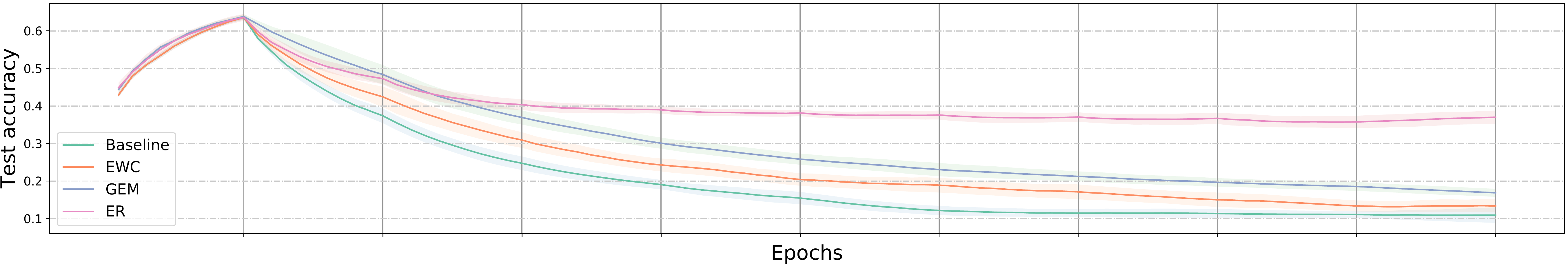}
  \caption{Kafnet}
  \label{cifar100:kaf}
\end{subfigure}

\caption{Evolution of accuracy on task 1 when training on CIFAR100.}
\label{cifar100:task1}
\end{figure}

Figs. \ref{cifar10:task1} and \ref{cifar100:task1} show how the score on the first task evolves during the training, respectively on CIFAR10 and CIFAR100. These images show how the proposed ER method is the best one on remembering the first task, with the exception of the Kafnet on CIFAR10, in which our method achieves better accuracy than GEM but lower remembering. The ER approach achieves better remembering on standard ANN trained on CIFAR10 given the less binding constraints imposed to the networks by the ER, w.r.t. GEM or EWC, because we regularize only the last layer of the network letting the network to adjust the internal weights as it sees fit, giving it more operating space while learning new tasks. This is the most important advantage over the other compared methods.

We complete the analysis in this section with a statistical analysis of the algorithms according to the procedure in \citet{demvsar2006statistical}. A Friedman rank test confirms that there are statistical significant differences with respect to the accuracy in Tab. \ref{table} (p-value of $0.01$). A successive set of Nemenyi post-hoc tests further confirms statistical significant differences between the proposed method, the baseline network, SI, and EWC.

\subsection{Impact of the memory size and computational time}
\label{subsec:impact_of_the_memory_size}

The previous section showed that the proposed ER method is able to achieve competiting results in terms of accuracy and remembering with a smaller external memory. In this section, we investigate how the performance of the methods varies when we increase or decrease the size of the memory.

\begin{figure}[t]
\centering
\begin{subfigure}{0.49\textwidth}
\centering
\resizebox{\textwidth}{!}{%
\begin{tikzpicture}
\begin{axis}[
    ybar,
    bar width=0.3cm,
    nodes near coords,
    enlargelimits=0.22,
    legend style={at={(0.5,-0.20)},
      anchor=north,legend columns=-1},
    symbolic x coords={10, 50, 100, 200, 300, 500},
    xtick=data,
        every node near coord/.append style={
                        anchor=west,
                        rotate=90
                },
    ]
    
    \addplot
    coordinates {
    (10, 0.84) (50,  0.85) (100,0.87) (200,0.88) (300,0.90) (500, 0.90)
    };
    
    \addplot
    coordinates {
    (10, 0.77) (50,  0.78) (100, 0.80) (200,0.81) (300,0.82) (500, 0.83)
    };
\end{axis}
\end{tikzpicture}}
\caption{GEM}
\end{subfigure}
\begin{subfigure}{0.49\textwidth}
\centering
\resizebox{\textwidth}{!}{%
\begin{tikzpicture}
\begin{axis}[
    ybar,
    bar width=0.3cm,
    nodes near coords,
    enlargelimits=0.22,
    legend style={at={(0.8,0.2)},
      anchor=north,legend columns=1},
    symbolic x coords={10, 50, 100, 200, 300, 500},
    xtick=data,
    every node near coord/.append style={
                        anchor=west,
                        rotate=90
                },
    ]
    
    \addplot
    coordinates {
    (10,  0.75) (50,  0.87) (100, 0.91) (200, 0.87) (300, 0.90) (500, 0.85)
    };
    
    \addplot
    coordinates {
    (10,  0.75) (50,  0.83) (100, 0.86) (200,  0.83) (300,0.86) (500, 0.82)
    };
    \legend{Remembering , Accuracy}

\end{axis}
\end{tikzpicture}}
\caption{Embedding}
\end{subfigure}
\caption{A comparison between GEM and ER when varying the memory size.}
\label{cifar10:mem_study}
\end{figure}
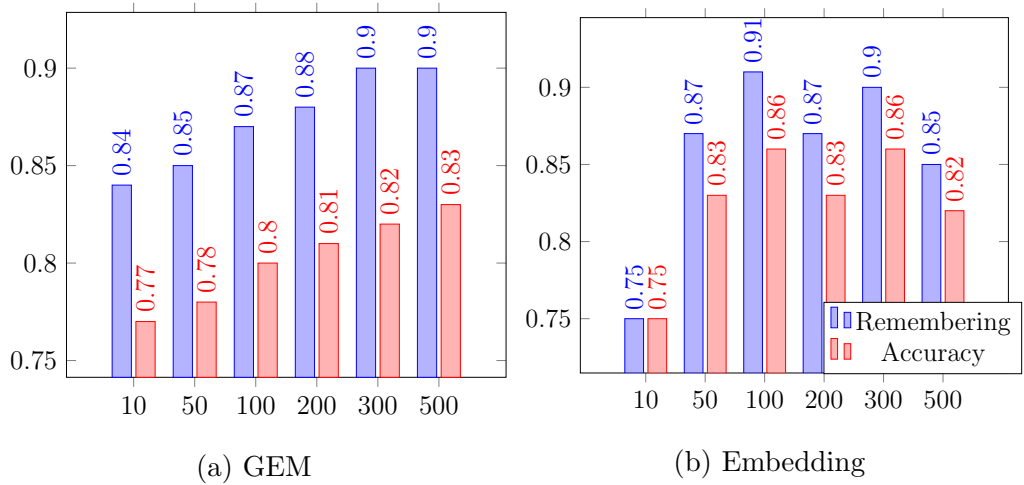

To this end, Fig. \ref{cifar10:mem_study} shows the accuracy and the remembering for both ER and GEM as a function of the memory associated to each task. GEM benefits from a bigger memory while ER achieves the best score around 100. 

\begin{figure}[H]
\centering
\begin{subfigure}{0.6\textwidth}
\centering
\resizebox{\textwidth}{!}{%
\begin{tikzpicture}
\begin{axis}[
    xbar,
    nodes near coords,
    legend style={at={(0.8,0.9)},
      anchor=north,legend columns=1},
    symbolic y coords={CIFAR10, CIFAR100, MNIST},
    enlarge x limits  = 0.3,
    enlarge y limits  = 0.3,
    ytick=data,
    xticklabels={},
    ]
    
    \addplot
    coordinates {
    (2266,CIFAR10) (4456,CIFAR100) (2040,MNIST) 
    };
    
    \addplot
    coordinates {
    (1050,CIFAR10) (1500,CIFAR100) (540,MNIST)
    };
    
        \addplot
    coordinates {
    (600,CIFAR10) (780,CIFAR100) (580,MNIST)
    };
    
    \addplot
    coordinates {
    (328,CIFAR10) (778,CIFAR100) (320,MNIST)
    };
    
    \legend{GEM , ER, EWC, SI}

\end{axis}
\end{tikzpicture}}
\end{subfigure}
\caption{Training time required by the approaches (in seconds). The times are averaged over the ANN experiments.}
\label{cifar10:computational_time}
\end{figure}
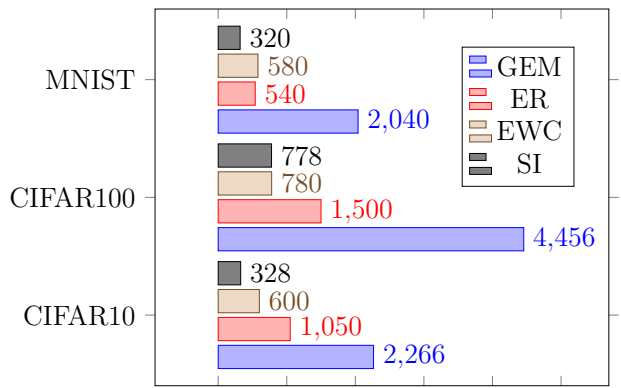

Training time required by the approaches is shown in Fig. \ref{cifar10:computational_time}. The image shows that, on CIFAR10, ER requires half of the time if compared to GEM, and also ER scales much better when the number of tasks increases (CIFAR100), making GEM not suitable.
We can conclude that ER achieves better scores using less memory and the training requires less time, not needing a QP optimization at every optimization step; in fact the QP optimization problem has a number of parameters equals to the number of tasks, making it slower when the number of tasks in the training grows. 

In the experiments about the time we considered also the memory size, but this value is not much impactful if compared to the QP optimization problem; in fact when the memory grows the required training time increases from 10 to 20 seconds, for both GEM and ER.


\subsection{Impact of the sampling strategy}
\label{subsec:impact_of_the_sampling_strategy}

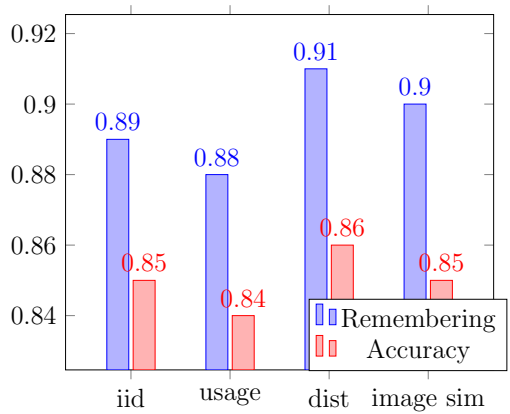
\begin{figure}[H]
\centering
\begin{subfigure}{0.49\textwidth}
\centering
\resizebox{\textwidth}{!}{%
\begin{tikzpicture}
\begin{axis}[
    ybar,
    bar width=10pt,
    nodes near coords,
    enlargelimits=0.22,
    legend style={at={(0.8,0.2)},
      anchor=north,legend columns=1},
    symbolic x coords={iid, usage, dist, image sim},
    xtick=data,
    ]
    
    \addplot
    coordinates {
    (iid,  0.89) (usage,  0.88) (dist, 0.91) (image sim, 0.90)
    };
    
    \addplot
    coordinates {
    (iid,  0.85) (usage,  0.84) (dist, 0.86) (image sim,  0.85)
    };
    \legend{Remembering , Accuracy}

\end{axis}
\end{tikzpicture}}
\end{subfigure}
\caption{A comparison between the sample size techniques used in ER.}
\label{cifar10:sample_study}
\end{figure}

Next, we investigate how the different sampling strategies described in Section \ref{subsec:selecting_and_updating_the_sampling_probabilities} impact the accuracy of the method. Fig. \ref{cifar10:sample_study} shows the scores obtained when using the different techniques. In general, the distance between the images used as weight performed well, but the real embedding distance achieves better scores.

The image shows also that a sampling approach based on the usage of the images leads to the worst results, since it doesn't provide us any information about the real importance of the images. The distance based sampling method is the best one because it provide useful information about the real distance from the past tasks and the current one, and these distances are adapted during the training, and does not requires any further computation (as opposed to image similarity).  

\subsection{Visualization of the embeddings}

\begin{figure}[H]
\begin{subfigure}{\textwidth}
  \centering
  \includegraphics[width=0.19\linewidth]{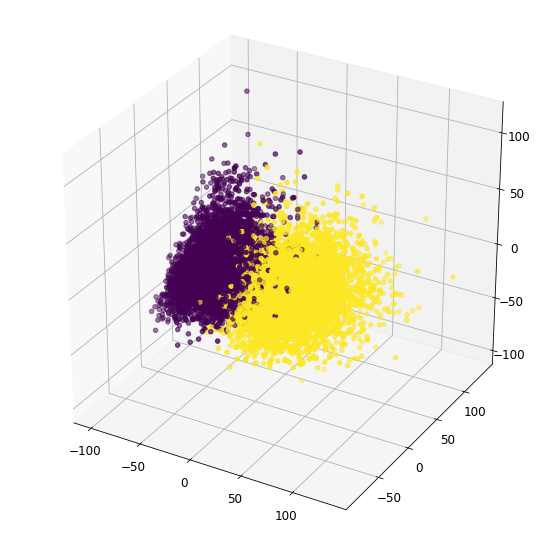}
  \hfill
  \includegraphics[width=0.19\linewidth]{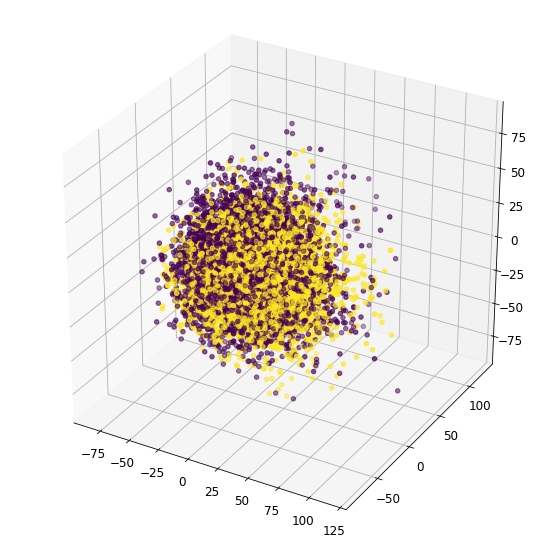}
    \hfill
  \includegraphics[width=0.19\linewidth]{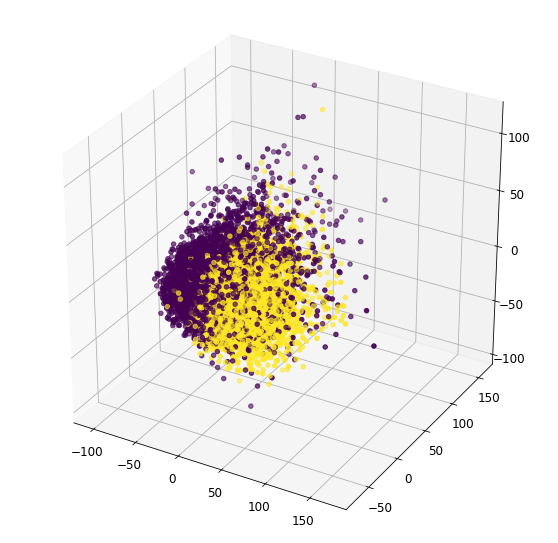}
    \hfill
  \includegraphics[width=0.19\linewidth]{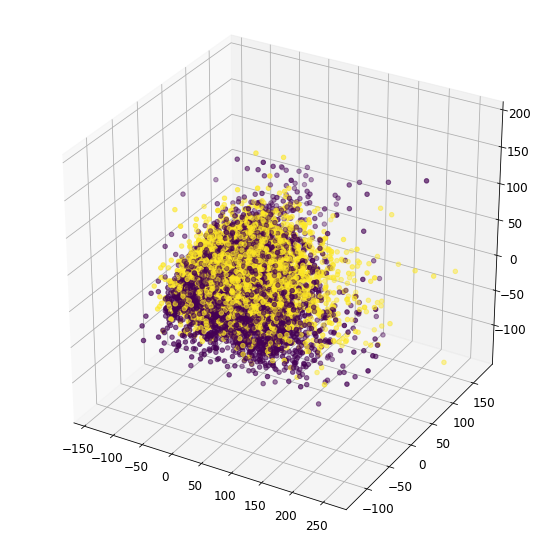}
    \hfill
  \includegraphics[width=0.19\linewidth]{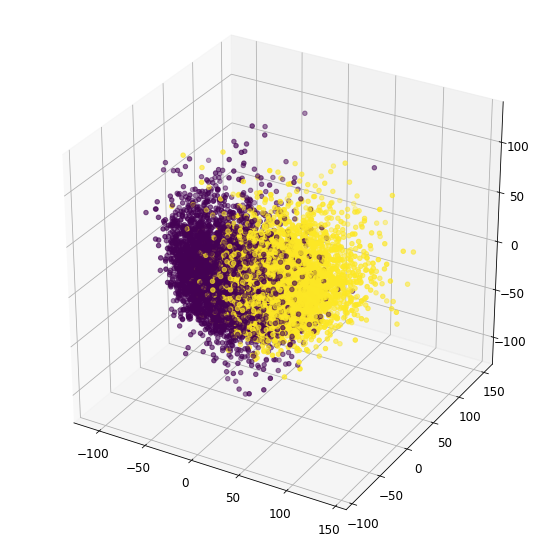}
  \caption{No approach to alleviate catastrophic forgetting}
\end{subfigure}

\begin{subfigure}{\textwidth}
  \centering
  \includegraphics[width=0.19\linewidth]{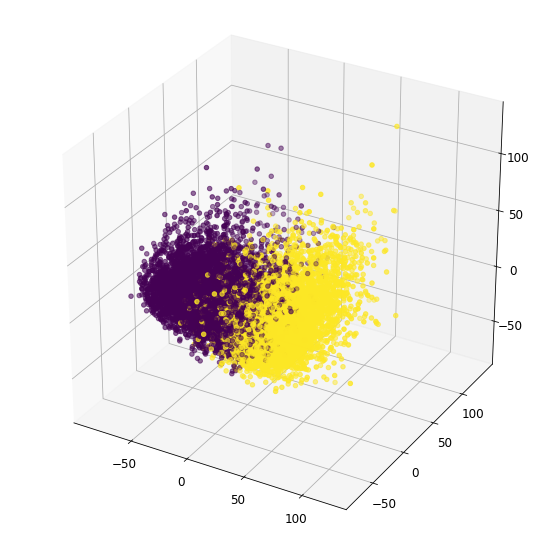}
  \hfill
  \includegraphics[width=0.19\linewidth]{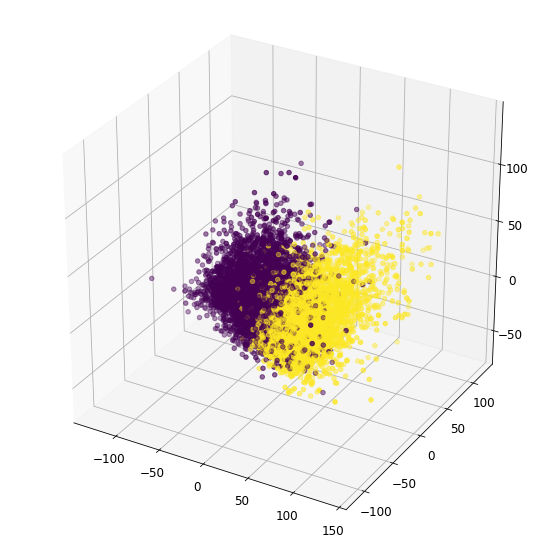}
    \hfill
  \includegraphics[width=0.19\linewidth]{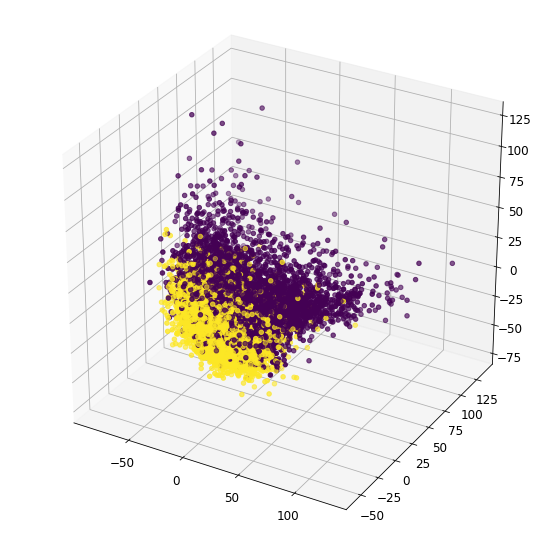}
    \hfill
  \includegraphics[width=0.19\linewidth]{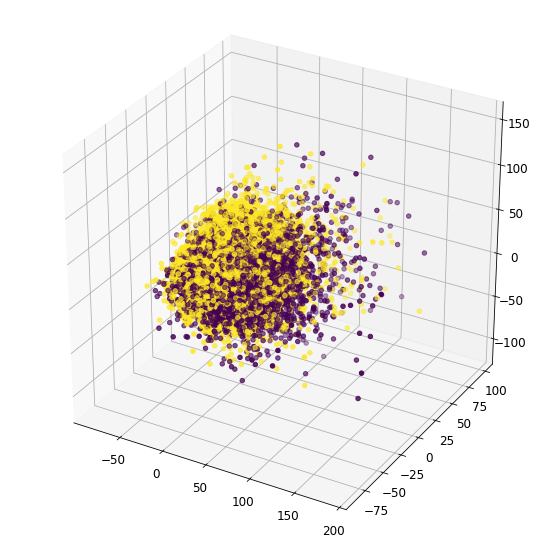}
    \hfill
  \includegraphics[width=0.19\linewidth]{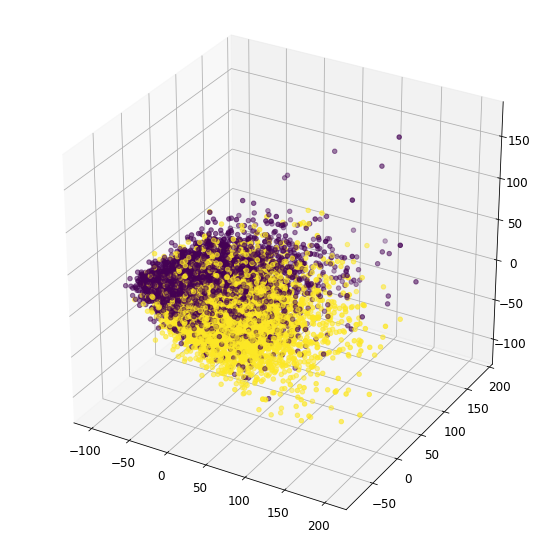}
  \caption{EWC}
\end{subfigure}

\begin{subfigure}{\textwidth}
  \centering
  \includegraphics[width=0.19\linewidth]{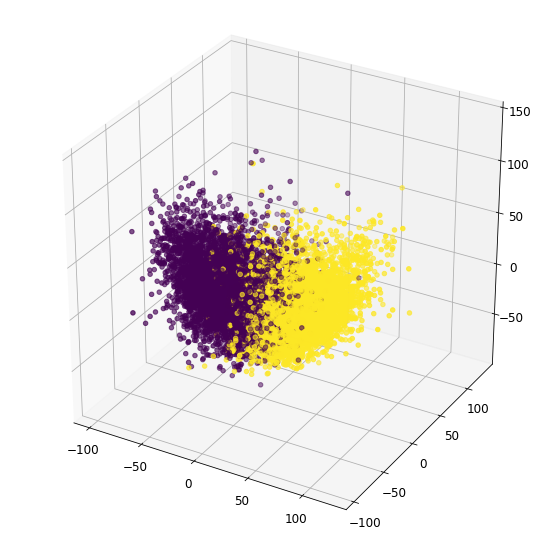}
  \hfill
  \includegraphics[width=0.19\linewidth]{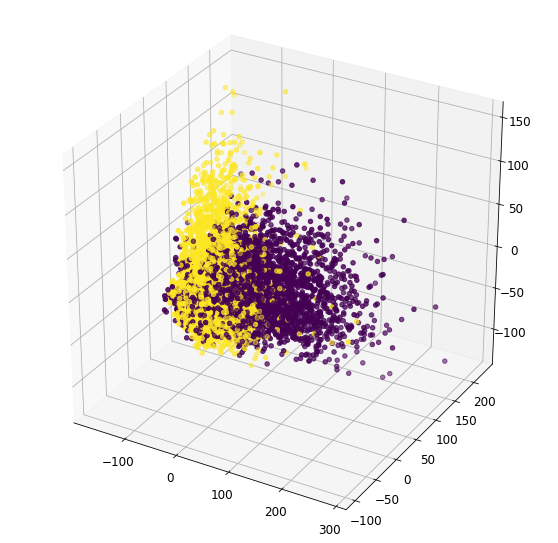}
    \hfill
  \includegraphics[width=0.19\linewidth]{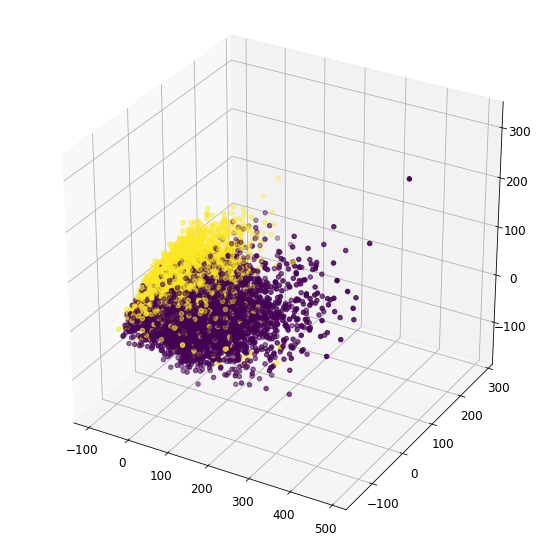}
    \hfill
  \includegraphics[width=0.19\linewidth]{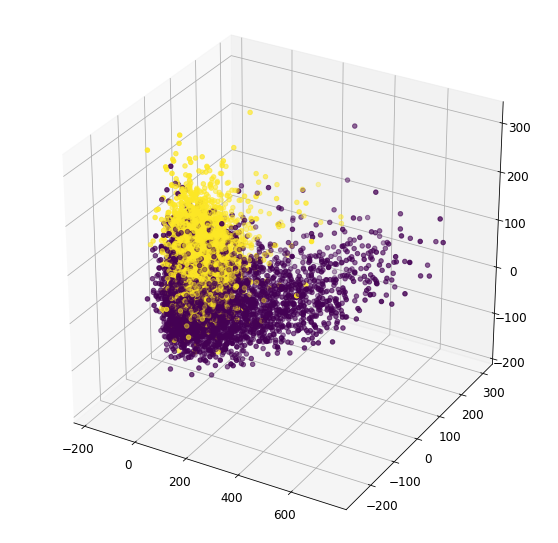}
    \hfill
  \includegraphics[width=0.19\linewidth]{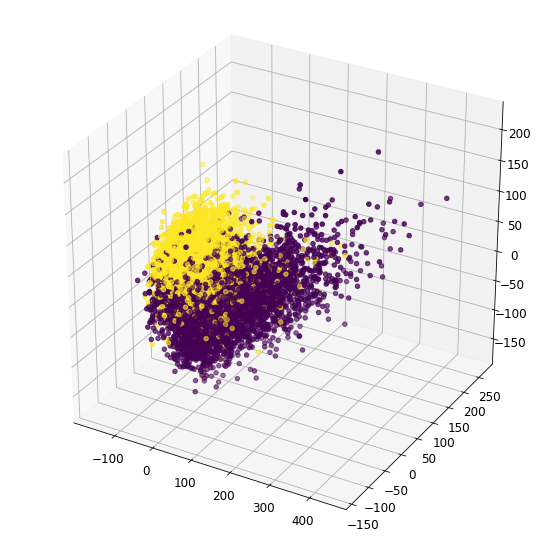}
  \caption{GEM}
\end{subfigure}

\begin{subfigure}{\textwidth}
  \centering
  \includegraphics[width=0.19\linewidth]{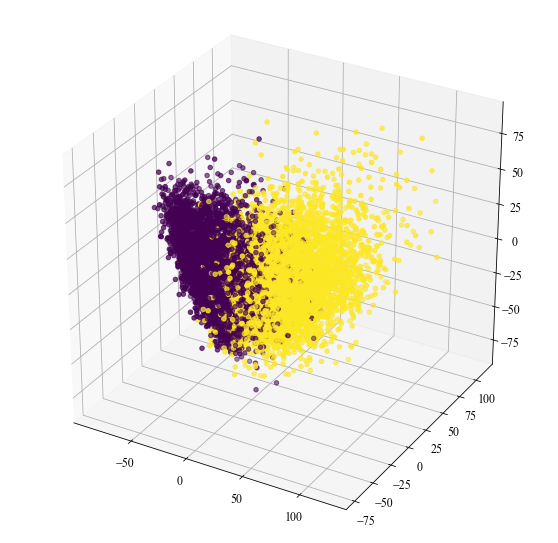}
  \hfill
  \includegraphics[width=0.19\linewidth]{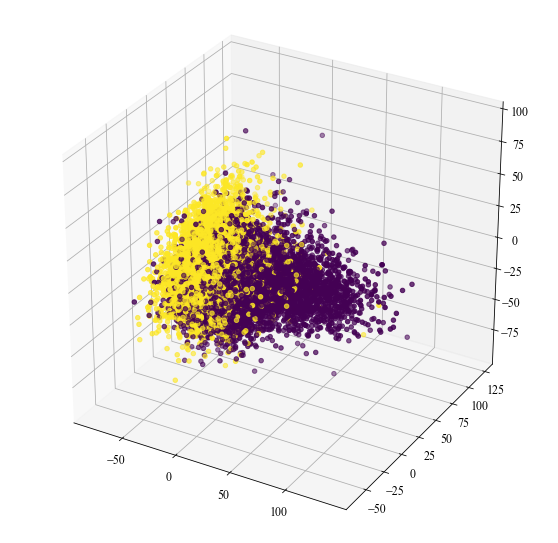}
    \hfill
  \includegraphics[width=0.19\linewidth]{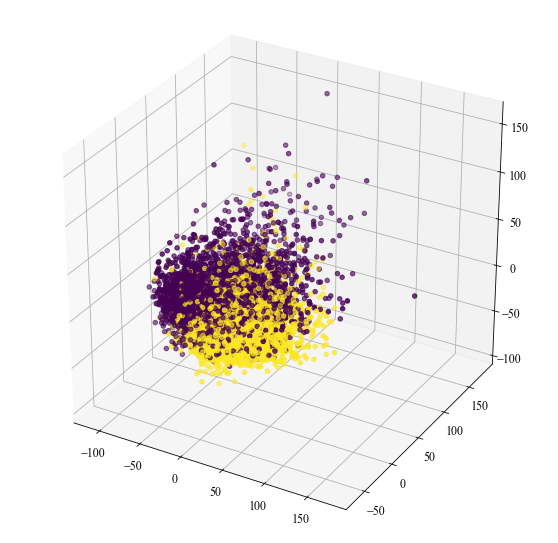}
    \hfill
  \includegraphics[width=0.19\linewidth]{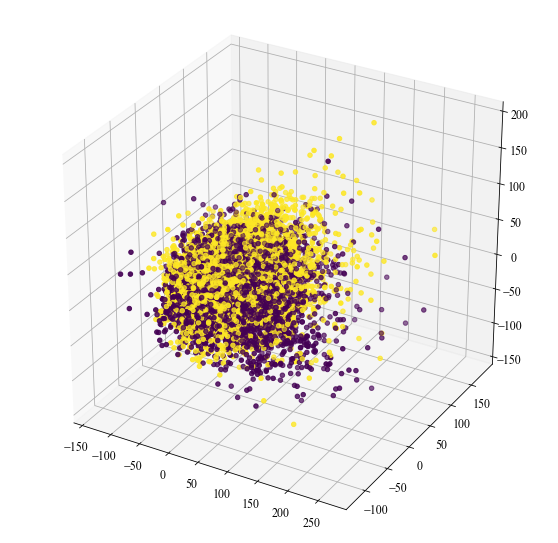}
    \hfill
  \includegraphics[width=0.19\linewidth]{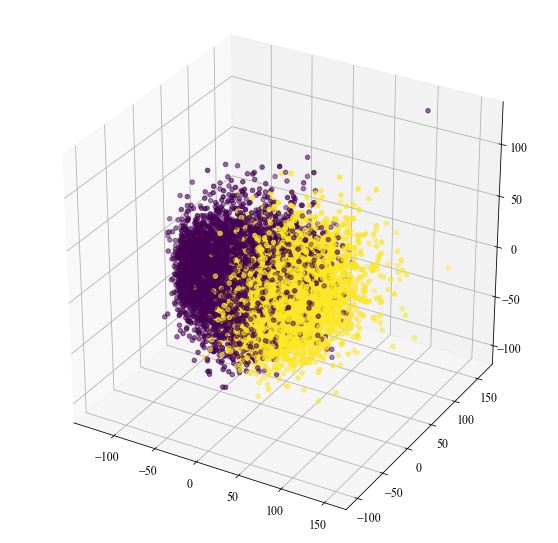}
  \caption{SI}
\end{subfigure}

\begin{subfigure}{\textwidth}
  \centering
  \includegraphics[width=0.19\linewidth]{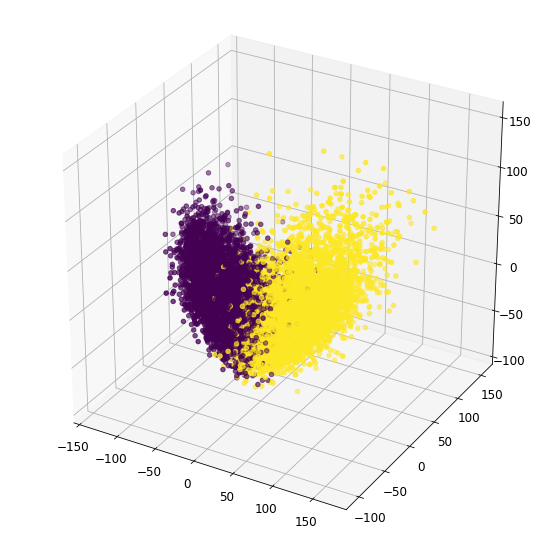}
  \hfill
  \includegraphics[width=0.19\linewidth]{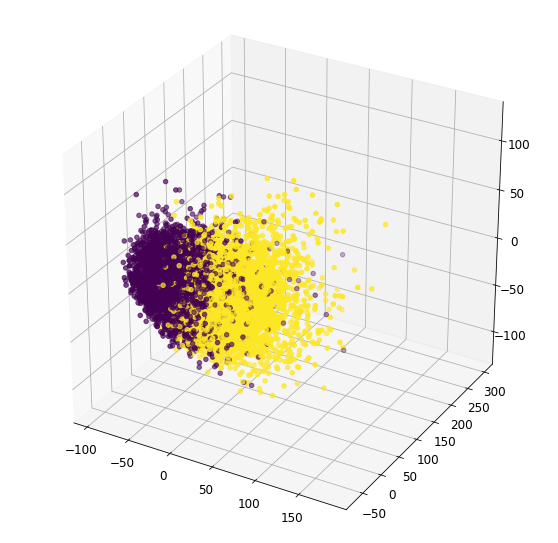}
    \hfill
  \includegraphics[width=0.19\linewidth]{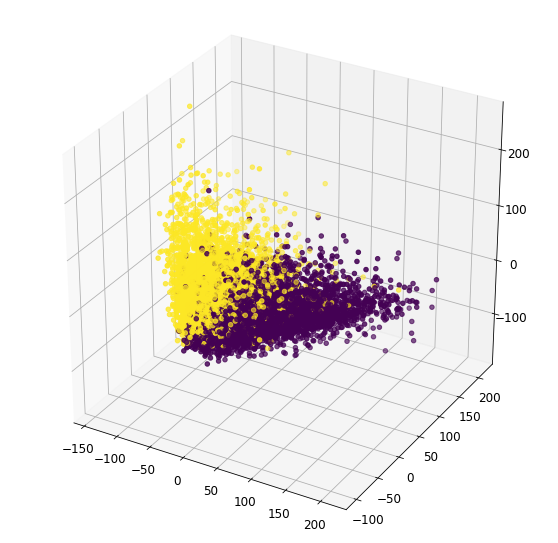}
    \hfill
  \includegraphics[width=0.19\linewidth]{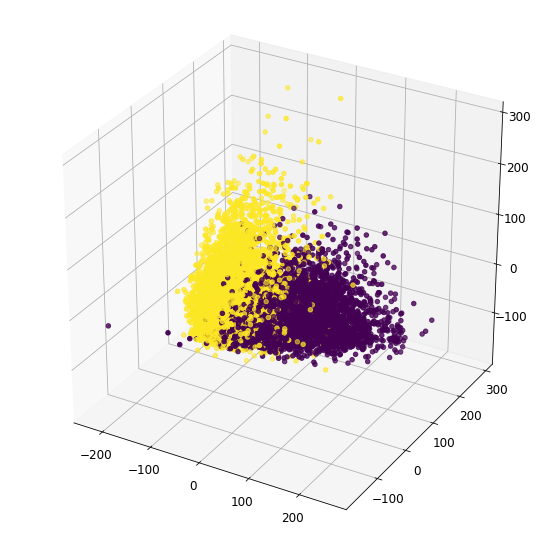}
    \hfill
  \includegraphics[width=0.19\linewidth]{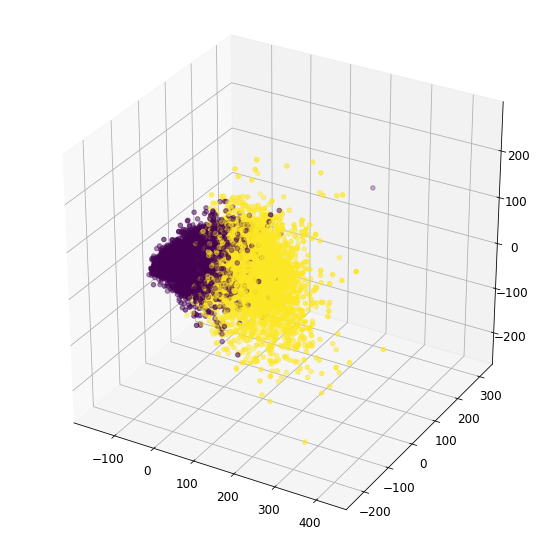}
  \caption{ER}
\end{subfigure}
\caption{How the feature space of the first CIFAR10 task changes from one task to another. Each column contains the feature space at the end of the training of a task from CIFAR10, from 1 to 5. Visualized using PCA algorithm.}
\label{cifar10:embeddings}
\end{figure}

Finally, it could be interesting to see how the features space changes during the training, to see if the intuitions behind the ER method are empirically confirmed. Figure \ref{cifar10:embeddings} shows that GEM is capable of keeping the embedding shape unchanged during the training, however the division between the classes become less clear in the end of the training. ER method rotates the points, since the distance used are the cosine dissimilarity and it constrains only the direction of the embeddings, but the division remains neat during all the training. 

We can now confirm that regularizing the embeddings is a promising approach to alleviate catastrophic forgetting in a CL multi-task scenario. 


\section{Conclusions}
\label{sec:conclusions}

In this paper, we proposed a new method for continual learning of neural networks called embedding regularization, which uses the features vectors extracted from the past tasks to regularize the training on the current task. Embedding has several advantageous characteristics, if compared to other state-of-the-art techniques such as EWC and GEM. First, the method is easier to implement, since it does not work directly with the parameters but let the network adapts itself based on the embeddings; the only constraint is that the network must be able to extract the information from the past tasks. 
Second, as shown by our experimental evaluation, the method requires very small memory footprint to works properly. The combination of the previous points leads also to a faster training and a better averaged accuracy. 

Our method suggests several interesting lines of further research, in which it is not necessary to operate directly on the network to alleviate catastrophic forgetting, but one can design intermediate auxiliary objectives describing the right constraints, such that the network can adapt itself during the training. An interesting line of research is trying to combine the strengths of the proposed ER method with the other state-of-the-art techniques known for CL, also beyond regularization approaches.

\section*{References}
\bibliographystyle{elsarticle-harv}
\bibliography{biblio}

\end{document}